\documentclass[conference]{IEEEtran}
\IEEEoverridecommandlockouts

\usepackage{cite}
\usepackage{amsmath,amssymb,amsfonts}
\usepackage{algorithmic}
\usepackage{graphicx}
\usepackage{textcomp}
\usepackage{xcolor}
\def\BibTeX{{\rm B\kern-.05em{\sc i\kern-.025em b}\kern-.08em
    T\kern-.1667em\lower.7ex\hbox{E}\kern-.125emX}}
\begin{document}

\title{Discrete Ricci Curvature on Protein Contact Graphs for Lightweight Fold Classification

}

\author{\IEEEauthorblockN{Jianru Shen}
\IEEEauthorblockA{
\textit{University of Montana}\\
Missoula, MT, USA \\
js258133@umconnect.umt.edu}}

\maketitle

\begin{abstract}
Protein fold classification can be approached via sequence-based
representations or structural descriptors, but direct comparisons between
lightweight handcrafted descriptors and pretrained protein language model
embeddings remain limited. We investigate discrete Ricci curvature on
C$\alpha$ contact graphs as a lightweight structural descriptor for fold
classification. Each protein domain is represented by a 22-dimensional fixed-length
feature derived from summary statistics and quantiles of Ollivier-Ricci
and Forman-Ricci edge curvature distributions. We evaluate on CATH top-10 Topology classification and on the ASTRAL
40\%-identity SCOPe top-10 Fold benchmark, comparing against geometry,
contact-graph statistics, persistent homology, and mean-pooled ESM-2
(150M) baselines.
On both datasets, lightweight structural descriptors substantially
outperform mean-pooled ESM-2 embeddings, with a larger performance gap
on the ASTRAL 40\% SCOPe benchmark. Ricci alone uses 22 dimensions, or 3.4\% of the ESM-2 baseline
dimensionality, and already outperforms mean-pooled ESM-2 on both
datasets. Combining Ricci with persistent homology yields the strongest
performance, achieving macro-F1 of 0.71 on CATH and 0.68 on SCOPe with a
112-dimensional feature vector. These results identify
a regime where lightweight interpretable graph descriptors offer a
practical alternative to pretrained protein language model embeddings.
\end{abstract}


\section{Introduction}

Protein fold classification organizes domains by their three-dimensional
structural arrangement and provides a foundation for function prediction,
evolutionary analysis, and structure-guided design. Hierarchical resources
such as CATH \cite{cath} and SCOPe \cite{scope} have become standard
references for this task. The Topology level of CATH and the Fold level of
SCOPe both correspond to fold-like organization and are widely used as
domain-level classification targets.

Existing representations for fold classification fall into two broad
families. Sequence-based features range from amino acid k-mer compositions
to embeddings from large pretrained protein language models such as
ESM-2 \cite{esm2}. Structural descriptors include geometric summaries,
contact graph statistics, persistent homology \cite{ph_protein_wei},
and related topological tools. Pretrained protein embeddings are powerful but require pretrained-model
inference and can be difficult to interpret. Lightweight
handcrafted structural descriptors remain attractive in settings where
training data is limited, structural inductive bias is desired, or
interpretability matters. Within this lightweight family, the role of
discrete Ricci curvature on protein contact graphs has received limited
systematic evaluation, and direct comparisons between handcrafted graph
descriptors and mean-pooled protein language model embeddings on small-data
fold classification remain rare.

We investigate discrete Ricci curvature \cite{ollivier, forman} as a
lightweight structural descriptor for protein fold classification. Each
domain is represented as a C$\alpha$ contact graph, on which we compute
both Ollivier-Ricci and Forman-Ricci curvature. The Ollivier formulation is
grounded in optimal transport and theoretically rich, while the Forman
formulation is combinatorial and substantially cheaper. We summarize each per-edge curvature distribution by its summary
statistics and quantiles, producing a 22-dimensional fixed-length
feature vector per protein. We
position Ricci curvature as a local discrete geometry descriptor that
complements multi-scale topological descriptors such as persistent
homology, and we evaluate it on both CATH and SCOPe to assess
cross-dataset robustness.

Our contributions are as follows.
\begin{itemize}
    \item We present a lightweight discrete Ricci curvature feature
    pipeline for protein fold classification using C$\alpha$ contact graphs,
    covering both Ollivier-Ricci and Forman-Ricci formulations.
    \item We empirically show that lightweight structural graph
    descriptors substantially outperform mean-pooled ESM-2 (150M)
    embeddings on CATH top-10 Topology classification and SCOPe top-10
    Fold classification, with a larger performance gap on the ASTRAL
    40\% SCOPe benchmark.
    \item We show that Ricci features combine complementarily with
    persistent homology, with their combination achieving the strongest
    performance among all evaluated descriptors on both datasets.
\end{itemize}

\begin{figure*}[t]
\centering
\includegraphics[width=0.95\textwidth]{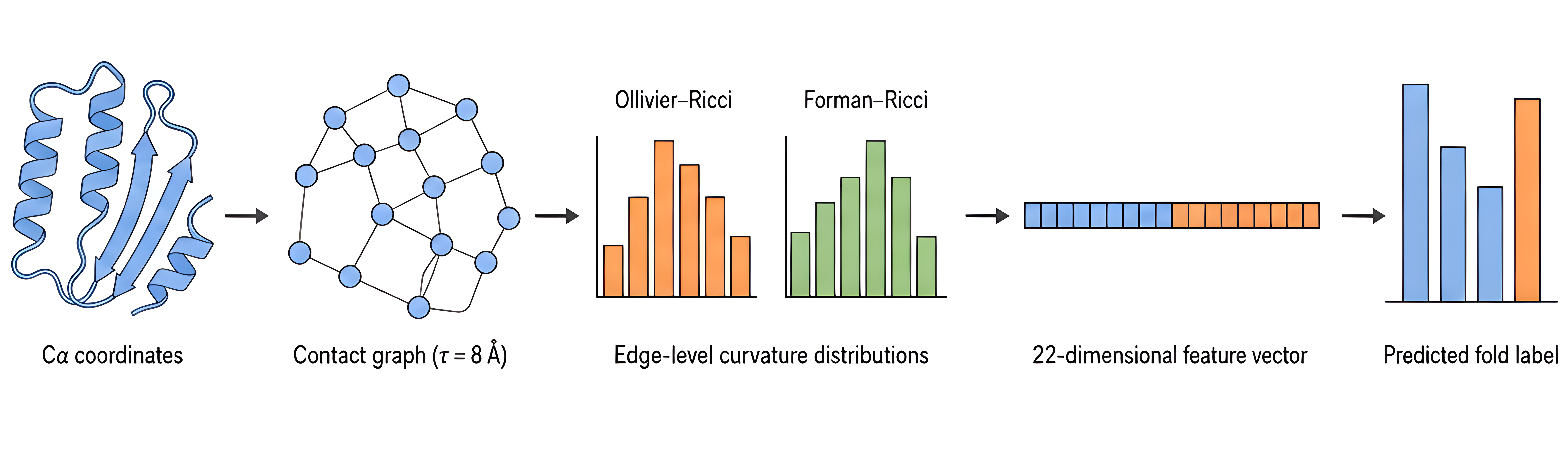}
\caption{Overall pipeline. A protein domain is represented as a C$\alpha$
contact graph at threshold $\tau$. Ollivier-Ricci and Forman-Ricci
curvature are computed on graph edges, and each per-edge curvature
distribution is represented by six statistics and five quantiles. The
resulting 22-dimensional fixed-length feature is fed to a classifier
for fold prediction.}
\label{fig:pipeline}
\end{figure*}

\section{Related Work}

\subsection{Protein Fold Classification}

Protein fold classification has long relied on sequence similarity
methods such as PSI-BLAST \cite{psiblast} and HMM-based profile
searches \cite{hmmer}. Structural alignment tools including DALI
\cite{dali} and TM-align \cite{tmalign} provide direct geometric
comparison and are commonly used to build and evaluate hierarchical
classifications like CATH \cite{cath} and SCOPe \cite{scope}. More
recently, pretrained protein language models such as ESM-2 \cite{esm2}
have enabled embedding-based approaches that exploit large-scale
sequence statistics. Our work differs by focusing on lightweight handcrafted graph
descriptors and providing a direct comparison with mean-pooled
language model embeddings under a common downstream evaluation
protocol.

\subsection{Persistent Homology in Biomolecular Analysis}

Persistent homology has been applied extensively to biomolecular
structure analysis. Cang and Wei \cite{cang_wei} introduced topological
deep learning for protein-ligand binding affinity prediction, and
subsequent work extended persistent homology to a range of biomolecular
characterization tasks \cite{ph_protein_wei}. These approaches
typically focus on per-task model design where persistent homology is
either the sole representation or one component of a learned model.
We instead use persistent homology as one of several lightweight
descriptors in a unified comparison, and we examine its complementarity
with curvature-based features rather than its standalone capability.

\subsection{Discrete Ricci Curvature on Networks}

Discrete Ricci curvature, introduced by Ollivier \cite{ollivier} via
optimal transport and by Forman \cite{forman} via combinatorial means,
has been used to analyze network structure across diverse domains
including community detection \cite{ricci_community} and biological
interaction networks \cite{sandhu}. Simplified formulations have made
these descriptors practical for large networks \cite{sreejith}.
Discrete Ricci curvature has also been used in biomolecular descriptor
design. Wee and Xia developed Ollivier persistent Ricci curvature and
Forman persistent Ricci curvature representations for protein-ligand
binding affinity prediction \cite{wee_oprc, wee_fprc}. Our work
addresses a different setting: domain-level fold classification from
C$\alpha$ contact graphs, with a unified comparison against persistent
homology and mean-pooled protein language model embeddings. To our
knowledge, discrete Ricci curvature has not previously been evaluated
as a fold-discriminative descriptor for protein domain classification.

\section{Method}

\subsection{Protein Structure Representation}

Each protein domain is represented as a contact graph derived from its
C$\alpha$ atom coordinates. Given a domain with $n$ residues, let
$\mathbf{r}_i \in \mathbb{R}^3$ denote the C$\alpha$ coordinate of
residue $i$. We construct an unweighted undirected simple graph
$G = (V, E)$ where $V = \{1, \ldots, n\}$ and an edge $\{i, j\} \in E$
is included whenever $i \neq j$ and
$\| \mathbf{r}_i - \mathbf{r}_j \|_2 \leq \tau$. We use
$\tau = 8\,\text{\AA}$ as the default threshold and study sensitivity
to this choice in Section~\ref{sec:experiments}. The same
representation pipeline is applied to both CATH and SCOPe domains.

\subsection{Ollivier-Ricci Curvature}

Ollivier-Ricci curvature \cite{ollivier} generalizes Ricci curvature to
discrete metric spaces by comparing the rate at which mass spreads from
neighboring points. For each node $x \in V$, we define the lazy random
walk distribution $\mu_x$ as
\begin{equation}
\mu_x(y) =
\begin{cases}
\alpha, & y = x \\
(1 - \alpha) / |\mathcal{N}(x)|, & y \in \mathcal{N}(x) \\
0, & \text{otherwise}
\end{cases}
\end{equation}
where $\mathcal{N}(x)$ denotes the set of neighbors of $x$,
$|\mathcal{N}(x)| = \deg(x)$, and $\alpha$ controls the lazy mass
retained at $x$. We fix $\alpha = 0.5$. The Ollivier-Ricci curvature
of an edge $\{x, y\}$ is
\begin{equation}
\kappa_O(x, y) = 1 - \frac{W_1(\mu_x, \mu_y)}{d(x, y)}
\end{equation}
where $W_1$ is the Wasserstein-1 distance under the graph shortest-path
metric. Since curvature is evaluated only on edges, $d(x, y) = 1$ for
every pair considered. We compute $W_1$ exactly using the POT
library \cite{pot}. Positive $\kappa_O$ indicates that random walks
from neighboring nodes overlap tightly, suggesting locally cluster-like
structure. Negative $\kappa_O$ indicates expansion, typical of
bridge-like edges between substructures.

\subsection{Forman-Ricci Curvature}

Forman-Ricci curvature \cite{forman} is a purely combinatorial discrete
curvature that requires no optimal transport. For an unweighted simple
graph, we use the standard formulation \cite{sreejith}
\begin{equation}
\kappa_F(x, y) = 4 - \deg(x) - \deg(y)
\end{equation}
which can be evaluated in $O(|E|)$ time. Forman curvature is therefore
substantially cheaper than Ollivier curvature, and we include it both as
an efficient alternative and as a complementary descriptor in our
ablation.

\subsection{Curvature Feature Vectorization}

Each protein produces two sets of per-edge curvature values, one from
Ollivier-Ricci curvature and one from Forman-Ricci curvature. We
summarize each set by 11 statistics: six summary statistics (mean,
standard deviation, skewness, kurtosis, minimum, and maximum) and five
quantiles at the 10th, 25th, 50th, 75th, and 90th percentiles. The two
11-dimensional summaries are concatenated into a 22-dimensional
fixed-length feature vector per protein.

\subsection{Classification Pipeline}

We use stratified 5-fold cross-validation with a fixed random seed for
all experiments. XGBoost \cite{xgboost} serves as the primary classifier
with 300 trees, maximum tree depth of 6, learning rate 0.1, subsample
0.8, and column subsample 0.8. Macro-F1 is reported as the primary metric to account for moderate
class size variation in SCOPe, where the smallest class contains 187
domains and the largest contains 300. Accuracy and
weighted-F1 are reported as secondary metrics. SVM and Random Forest
are evaluated as consistency checks and discussed in
Section~\ref{sec:experiments}.

\begin{figure*}[t]
\centering
\includegraphics[width=0.85\textwidth]{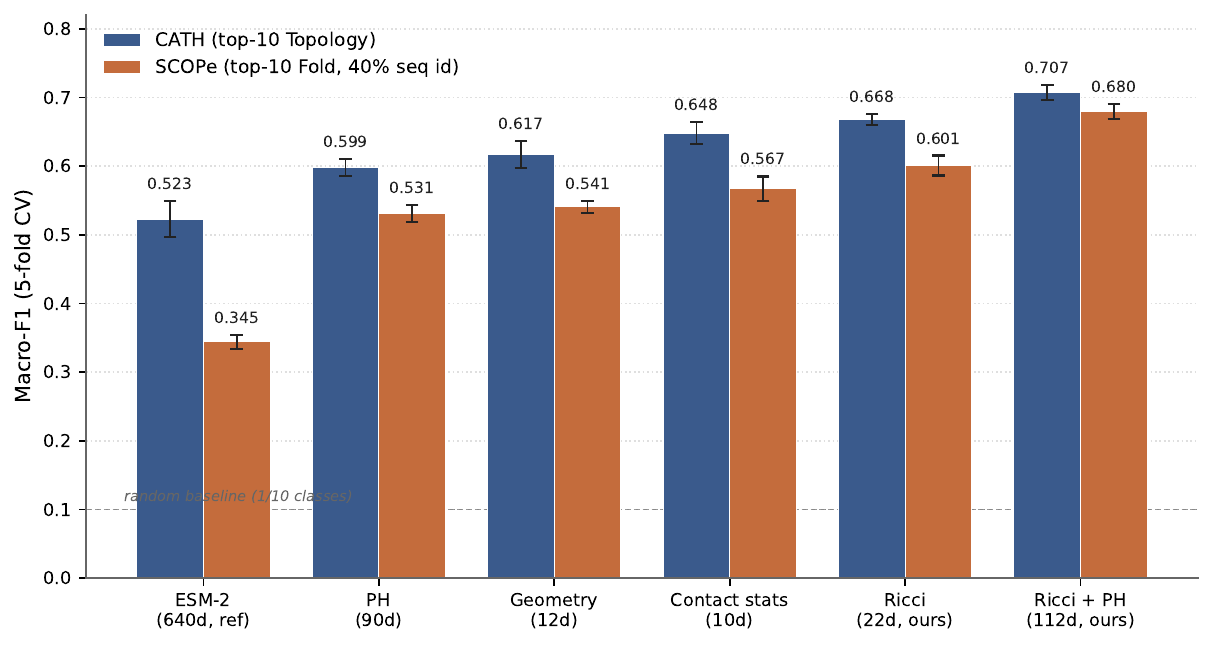}
\caption{Macro-F1 comparison across six feature groups on CATH (top-10
Topology) and SCOPe (top-10 Fold, ASTRAL 40\% sequence identity). Error
bars indicate one standard deviation across 5 folds. Random baseline
corresponds to 1/10 for ten-class classification. Lightweight structural descriptors substantially outperform mean-pooled
ESM-2 on both datasets, with a larger gap on the ASTRAL 40\% SCOPe
benchmark.}
\label{fig:cross_dataset}
\end{figure*}

\begin{figure*}[t]
\centering
\includegraphics[width=0.7\textwidth]{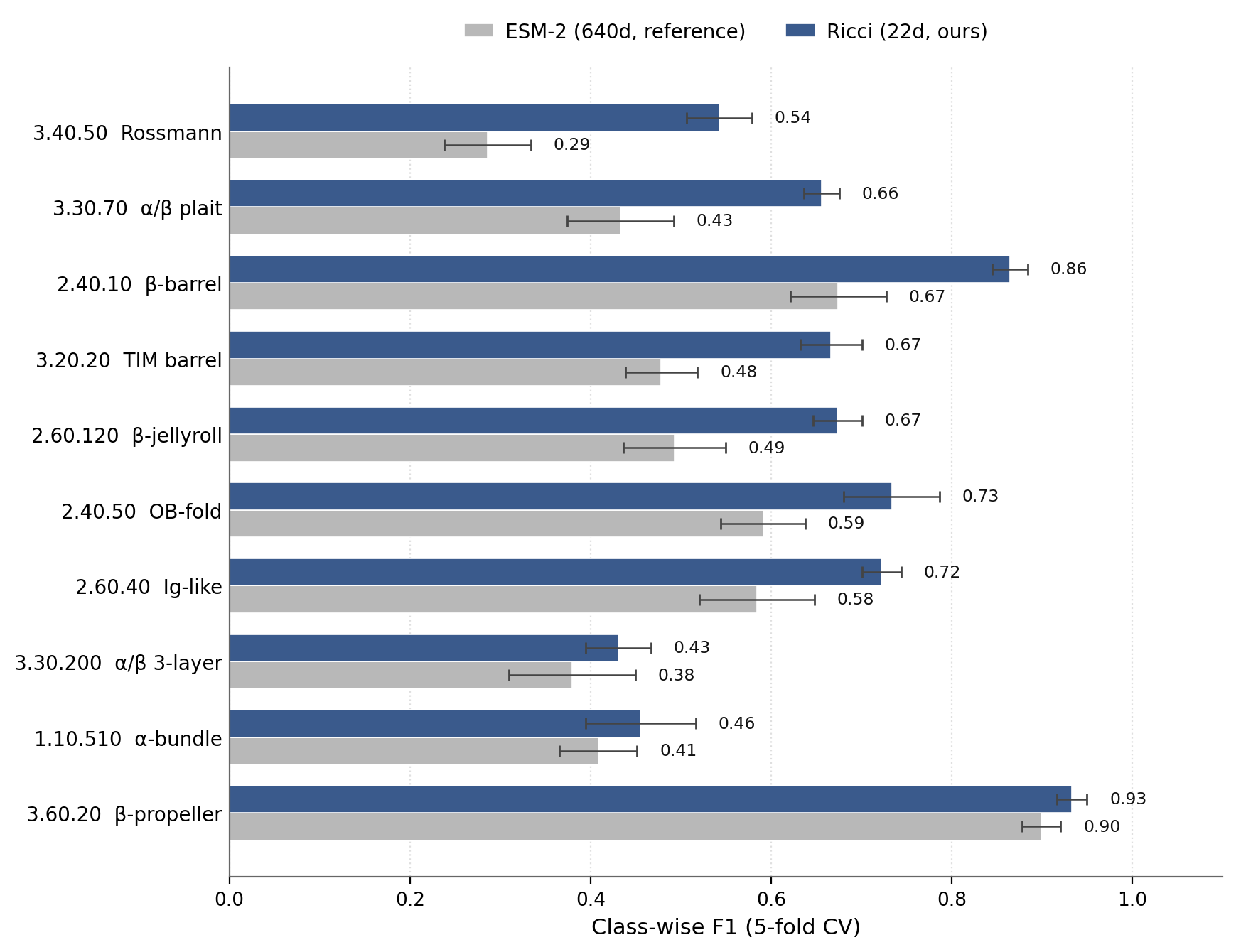}
\caption{Class-wise F1 breakdown on CATH for Ricci (22-dim, ours) and
ESM-2 (640-dim, reference), averaged over five cross-validation splits.
Structural fold classes are sorted by the Ricci-versus-ESM-2 gap,
largest at top. Ricci outperforms mean-pooled ESM-2 on every class,
with the largest absolute gap on the Rossmann fold (3.40.50, $+0.25$)
and the smallest on the $\beta$-propeller fold (3.60.20, $+0.03$).
Error bars indicate one standard deviation across five splits.}
\label{fig:per_class}
\end{figure*}

\section{Experiments}
\label{sec:experiments}

\subsection{Datasets}

We evaluate on two independently curated structural classification
resources. CATH \cite{cath} version 4.3 serves as the primary resource:
we select the top 10 most populated Topology classes and subsample each
class to between 200 and 300 domains, yielding 2998 domains in total
after C$\alpha$ extraction. SCOPe \cite{scope} version 2.08 is used for
cross-dataset robustness validation. We use the ASTRAL subset filtered at
40\% sequence identity \cite{astral}, select the top 10 most populated
Fold classes, and subsample analogously, yielding 2641 domains. All
experiments use stratified 5-fold cross-validation with a fixed random
seed (42) for reproducibility. The ASTRAL 40\% sequence-identity subset provides an additional
redundancy-controlled evaluation setting with reduced reliance on
sequence similarity.

\subsection{Feature Groups and Baselines}

We evaluate six feature groups summarized in Table~\ref{tab:features}.
Five are lightweight handcrafted structural descriptors derived from
C$\alpha$ contact graphs or point clouds. ESM-2 is included as a
reference pretrained embedding baseline, not as a head-to-head
competitor, since our focus is on lightweight descriptors that operate
under a fundamentally different compute regime. For ESM-2 we use the
t30 150M parameter variant \cite{esm2} with frozen weights and
mean-pooled final-layer representations.

The geometry baseline (12 dims) consists of residue count, radius of
gyration, diameter, and seven summary statistics of pairwise
C$\alpha$-C$\alpha$ distances (mean, standard deviation, minimum,
median, 25th and 75th percentiles, maximum), together with a
sphericity proxy ($R_g$ divided by diameter) and $\log$ of residue
count. The contact graph statistics baseline (10 dims) consists of mean,
standard deviation, minimum, and maximum of node degree, edge density,
mean local clustering coefficient, global transitivity, inverse edge
density, degree assortativity, and the normalized triangle count.

The persistent homology baseline is computed on the C$\alpha$ point
cloud of each domain using a Vietoris-Rips filtration with maximum
edge length 20\,\AA, implemented via GUDHI \cite{gudhi}. We extract
both 0-dimensional and 1-dimensional persistence diagrams, and
vectorize each by a 20-bin Betti curve and a $5 \times 5$ persistence
image \cite{persim}, yielding $2 \times (20 + 25) = 90$ dimensions per
protein. All
features, when standardized, are fitted on the training fold only and
applied to the held-out fold.

\begin{table}[t]
\caption{Feature groups evaluated in this work.}
\label{tab:features}
\centering
\begin{tabular}{lcc}
\hline
\textbf{Feature group} & \textbf{Dim} & \textbf{Type} \\
\hline
Geometry & 12 & lightweight \\
Contact graph statistics & 10 & lightweight \\
Persistent homology (H0+H1) & 90 & lightweight \\
Ricci (Ollivier + Forman) & 22 & lightweight (ours) \\
Ricci + PH & 112 & lightweight (ours) \\
ESM-2 (t30 150M, mean-pooled) & 640 & reference \\
\hline
\end{tabular}
\end{table}

\subsection{Main Comparison on CATH}

Table~\ref{tab:cath_main} reports macro-F1, accuracy, and weighted-F1 on
CATH top-10 Topology classification using XGBoost under 5-fold
cross-validation. Lightweight structural descriptors substantially
outperform mean-pooled ESM-2. Ricci alone (22 dimensions) achieves
macro-F1 of 0.668, exceeding ESM-2 (640 dimensions) by 0.145 absolute.
Combining Ricci with persistent homology yields macro-F1 of 0.707, the
strongest result among all evaluated descriptors. Additional checks
with an RBF-kernel SVM and a Random Forest produced qualitatively
similar rankings across feature groups.

\begin{table}[t]
\caption{CATH main results. XGBoost, 5-fold CV. Macro-F1 is reported as mean
$\pm$ standard deviation; accuracy and weighted-F1 are reported as
means. Best lightweight result in bold.}
\label{tab:cath_main}
\centering
\begin{tabular}{lcccc}
\hline
\textbf{Feature group} & \textbf{Dim} & \textbf{Macro-F1} & \textbf{Acc.} & \textbf{W-F1} \\
\hline
Geometry & 12 & 0.617 $\pm$ 0.019 & 0.618 & 0.617 \\
Contact stats & 10 & 0.648 $\pm$ 0.016 & 0.646 & 0.648 \\
PH & 90 & 0.599 $\pm$ 0.012 & 0.598 & 0.599 \\
Ricci & 22 & 0.668 $\pm$ 0.008 & 0.667 & 0.668 \\
\textbf{Ricci + PH} & 112 & \textbf{0.707 $\pm$ 0.011} & \textbf{0.706} & \textbf{0.707} \\
\hline
ESM-2 (reference) & 640 & 0.523 $\pm$ 0.027 & 0.521 & 0.523 \\
\hline
\end{tabular}
\end{table}

\subsection{Ricci Feature Ablation}

We dissect which components of the Ricci feature drive performance.
Table~\ref{tab:ricci_ablation} reports macro-F1 on CATH for five
configurations: Ollivier and Forman full summaries individually, their
concatenated full summary, a histogram-only variant, and the final
statistics-plus-quantiles variant. Ollivier and Forman provide
complementary information, with their combination outperforming either
alone. Among vectorization strategies, the 22-dimensional
statistics-plus-quantiles summary yields the best performance. We
adopt this configuration in all other experiments.

\begin{table}[t]
\caption{Ricci feature ablation on CATH. XGBoost, 5-fold CV. A full summary contains 6 statistics, 5 quantiles, and a 20-bin histogram per curvature. The final 22-dim feature uses statistics and quantiles from both curvatures.}
\label{tab:ricci_ablation}
\centering
\begin{tabular}{lcc}
\hline
\textbf{Configuration} & \textbf{Dim} & \textbf{Macro-F1} \\
\hline
Ollivier, full summary & 31 & 0.577 $\pm$ 0.020 \\
Forman, full summary & 31 & 0.550 $\pm$ 0.028 \\
Ollivier + Forman, full summary & 62 & 0.652 $\pm$ 0.014 \\
Ollivier + Forman, histogram only & 40 & 0.551 $\pm$ 0.016 \\
\textbf{Ollivier + Forman, statistics + quantiles} & \textbf{22} & \textbf{0.668 $\pm$ 0.008} \\
\hline
\end{tabular}
\end{table}

\subsection{Threshold Sensitivity and Compute Cost}

Table~\ref{tab:threshold} examines sensitivity to the contact graph
distance threshold $\tau$, with Ricci macro-F1 evaluated at
$\tau \in \{6, 8, 10, 12\}\,\text{\AA}$. All four thresholds yield
macro-F1 within 0.035 of one another, and the differences between
$\tau = 8\,\text{\AA}$ and $\tau = 12\,\text{\AA}$ fall within one
standard deviation across folds. Performance is therefore stable to
threshold choice within this range, and we adopt the standard
$\tau = 8\,\text{\AA}$ in all other experiments.

Table~\ref{tab:timing} reports wall-clock feature extraction time per
protein on a single Apple Silicon laptop. Forman-Ricci is over an order
of magnitude faster than Ollivier-Ricci or ESM-2 while remaining within
the same lightweight regime. The full Ricci + PH pipeline requires 231 ms per protein, in the same
order of magnitude as mean-pooled ESM-2 inference (152 ms), while not
depending on a pretrained model. Here, lightweight refers to compact handcrafted representations that require no pretrained model, rather than uniformly lower wall-clock time for every configuration.

\begin{table}[t]
\caption{Threshold sensitivity. Ricci (22-dim summary statistics + quantiles),
XGBoost, 5-fold CV on CATH.}
\label{tab:threshold}
\centering
\begin{tabular}{ccc}
\hline
$\boldsymbol{\tau}$ \textbf{(\AA)} & \textbf{Mean edges/protein} & \textbf{Macro-F1} \\
\hline
6 & 787 & 0.647 $\pm$ 0.012 \\
\textbf{8} & 1461 & \textbf{0.668 $\pm$ 0.008} \\
10 & 2650 & 0.639 $\pm$ 0.017 \\
12 & 4361 & 0.672 $\pm$ 0.015 \\
\hline
\end{tabular}
\end{table}

\begin{table}[t]
\caption{Compute cost: wall-clock per-protein feature extraction time on
a M4 MacBook (single-threaded; ESM-2 uses Apple MPS). Ricci + PH
combined cost is the sum of Ollivier-Ricci, Forman-Ricci, and PH.}
\label{tab:timing}
\centering
\begin{tabular}{lcc}
\hline
\textbf{Descriptor} & \textbf{ms/protein} & \textbf{Pretrained} \\
\hline
Geometry & 1.16 & No \\
Contact stats & 13.78 & No \\
Forman-Ricci & 6.69 & No \\
PH (H0+H1) & 74.24 & No \\
Ollivier-Ricci & 149.61 & No \\
Ricci + PH (estimated sum) & 230.54 & No \\
ESM-2 (150M, MPS) & 151.76 & Yes \\
\hline
\end{tabular}
\end{table}

\subsection{Per-Class Analysis on CATH}

To examine whether the advantage of lightweight descriptors holds
uniformly across fold classes, we report class-wise F1 scores for Ricci
and ESM-2 in Figure~\ref{fig:per_class}, averaged over five
cross-validation splits. Ricci outperforms mean-pooled ESM-2 on every
one of the ten fold classes. The gap is largest on the Rossmann fold
($+0.25$) and on the TIM barrel and $\alpha/\beta$ plait folds, where
ESM-2 falls below F1 of 0.50 while Ricci remains above 0.65. The gap
narrows on folds where both methods perform strongly, such as the
$\beta$-propeller fold ($+0.03$), and on folds where both methods
struggle, such as the $\alpha/\beta$ 3-layer fold ($+0.05$). Across all
classes, the class-wise pattern is consistent with the aggregate
result: lightweight discrete graph descriptors provide a uniform
advantage over mean-pooled protein language model embeddings on this
task, with the advantage being most pronounced on folds where
structural geometry, rather than sequence-level pattern, is the
dominant discriminator.

\subsection{Cross-Dataset Robustness on SCOPe}

To assess robustness, we apply the same pipeline to SCOPe top-10 Fold
classification on the 40\% sequence identity subset.
Table~\ref{tab:scope_main} reports the results. The ranking of feature
groups is preserved across both datasets: ESM-2 remains the weakest,
PH and Geometry are intermediate, and Ricci + PH achieves the best
macro-F1 on both CATH (0.707) and SCOPe (0.680). The cross-dataset
performance gap is smallest for Ricci + PH (0.027) and largest for
ESM-2 (0.178). The larger ESM-2 performance drop on SCOPe is consistent with the
hypothesis that mean-pooled language model embeddings rely more on
sequence-level patterns than structural descriptors do. However, because
CATH and SCOPe differ in curation, label hierarchy, and class
composition in addition to sequence identity threshold, this
cross-resource comparison should not be read as isolating the causal
effect of sequence identity filtering alone. Figure~\ref{fig:cross_dataset} visualizes both datasets side by
side, highlighting the consistency of the ranking across CATH and SCOPe.

\begin{table}[t]
\caption{\mbox{SCOPe} main results. Top-10 Fold classification on the
ASTRAL 40\% sequence identity subset. XGBoost, 5-fold CV.}
\label{tab:scope_main}
\centering
\begin{tabular}{lcc}
\hline
\textbf{Feature group} & \textbf{Dim} & \textbf{Macro-F1} \\
\hline
Geometry & 12 & 0.541 $\pm$ 0.009 \\
Contact stats & 10 & 0.567 $\pm$ 0.018 \\
PH & 90 & 0.531 $\pm$ 0.012 \\
Ricci & 22 & 0.601 $\pm$ 0.015 \\
\textbf{Ricci + PH} & 112 & \textbf{0.680 $\pm$ 0.011} \\
\hline
ESM-2 (reference) & 640 & 0.345 $\pm$ 0.010 \\
\hline
\end{tabular}
\end{table}

\section{Discussion}

The pattern observed across CATH and SCOPe admits a simple
interpretation. Protein fold classification is fundamentally a
structural task, and lightweight descriptors that encode geometric or
topological structure directly inherit the right inductive bias.
Mean-pooled embeddings from a sequence-only pretrained model do not
explicitly encode three-dimensional structure, leaving only an
aggregate signal that may correlate with fold through sequence
similarity. The 40\% sequence identity filter applied to SCOPe likely
contributes to this pattern by reducing sequence-level cues that
mean-pooled embeddings depend on. We refrain from a stronger causal
statement because CATH and SCOPe also differ in curation and label
organization, and the present comparison cannot disentangle these
factors from sequence identity filtering alone.

Among lightweight descriptors, Ricci curvature and persistent homology
capture distinct aspects of structure. Ricci summarizes local discrete
geometry at the edge level, while persistent homology summarizes
multi-scale topological features across a filtration. The consistent
improvement of Ricci + PH over either alone, on both datasets, suggests
that these two views are mathematically and empirically complementary.
Forman-Ricci provides a cheaper but somewhat weaker standalone descriptor
than Ollivier-Ricci, while running over twenty times faster and
contributing useful complementary information when combined.

These results do not argue against pretrained protein language models.
They identify an operating regime, namely small-data structural
classification, including sequence-redundancy-controlled settings, in
which lightweight interpretable descriptors are not merely competitive
with mean-pooled language model embeddings but substantially stronger.
Such regimes arise naturally in domain-specific analyses where
pretrained-model deployment is undesirable or unavailable, or where
interpretability is desired.

Several limitations qualify these findings. Each domain is represented
only by C$\alpha$ coordinates, omitting side-chain geometry and
atom-level chemistry. Our top-10 fold subsets do not span the full
diversity of CATH or SCOPe, and smaller or less represented fold
classes are not evaluated. The CATH experiment does not apply the same
sequence identity redundancy control as the ASTRAL 40\% SCOPe
benchmark, and a within-resource evaluation across matched identity
thresholds would provide a cleaner assessment of robustness to
sequence redundancy. Finally, we evaluate a single ESM-2 variant with
mean pooling; larger models, per-residue pooling, attention-based
aggregation, or fine-tuning may yield different conclusions and
warrant future investigation.

\section{Conclusion}

We presented a lightweight protein fold classification pipeline based on
discrete Ricci curvature of C$\alpha$ contact graphs, summarized by a
22-dimensional distribution-summary feature vector. Across CATH top-10
Topology classification and SCOPe top-10 Fold classification, Ricci
alone uses 3.4\% of the ESM-2 feature dimensionality and already
substantially outperforms mean-pooled ESM-2 embeddings. Combining Ricci
with persistent homology in a 112-dimensional feature vector achieves
macro-F1 of 0.71 on CATH and 0.68 on SCOPe. These results identify a
regime where lightweight handcrafted graph descriptors offer a
practical and interpretable alternative to pretrained protein language
model embeddings.


\begin{thebibliography}{00}
\bibitem{cath} I. Sillitoe, N. Bordin, N. Dawson, V. P. Waman, P. Ashford, H. M. Scholes, C. S. M. Pang, L. Woodridge, C. Rauer, N. Sen, M. Abbasian, S. Le Cornu, S. D. Lam, K. Berka, I. H. Varekova, R. Svobodova, J. Lees, and C. A. Orengo, ``CATH: increased structural coverage of functional space,'' \textit{Nucleic Acids Res.}, vol. 49, no. D1, pp. D266--D273, Jan. 2021.

\bibitem{scope} N. K. Fox, S. E. Brenner, and J.-M. Chandonia, ``SCOPe: Structural Classification of Proteins---extended, integrating SCOP and ASTRAL data and classification of new structures,'' \textit{Nucleic Acids Res.}, vol. 42, pp. D304--D309, Jan. 2014.

\bibitem{astral} J.-M. Chandonia, N. K. Fox, and S. E. Brenner, ``SCOPe: classification of large macromolecular structures in the structural classification of proteins-extended database,'' \textit{Nucleic Acids Res.}, vol. 47, no. D1, pp. D475--D481, Jan. 2019.

\bibitem{esm2} Z. Lin, H. Akin, R. Rao, B. Hie, Z. Zhu, W. Lu, N. Smetanin, R. Verkuil, O. Kabeli, Y. Shmueli, A. dos Santos Costa, M. Fazel-Zarandi, T. Sercu, S. Candido, and A. Rives, ``Evolutionary-scale prediction of atomic-level protein structure with a language model,'' \textit{Science}, vol. 379, no. 6637, pp. 1123--1130, Mar. 2023.

\bibitem{psiblast} S. F. Altschul, T. L. Madden, A. A. Schaffer, J. Zhang, Z. Zhang, W. Miller, and D. J. Lipman, ``Gapped BLAST and PSI-BLAST: a new generation of protein database search programs,'' \textit{Nucleic Acids Res.}, vol. 25, no. 17, pp. 3389--3402, Sep. 1997.

\bibitem{hmmer} S. R. Eddy, ``Accelerated profile HMM searches,'' \textit{PLoS Comput. Biol.}, vol. 7, no. 10, p. e1002195, Oct. 2011.

\bibitem{dali} L. Holm and C. Sander, ``Protein structure comparison by alignment of distance matrices,'' \textit{J. Mol. Biol.}, vol. 233, no. 1, pp. 123--138, Sep. 1993.

\bibitem{tmalign} Y. Zhang and J. Skolnick, ``TM-align: a protein structure alignment algorithm based on the TM-score,'' \textit{Nucleic Acids Res.}, vol. 33, no. 7, pp. 2302--2309, Apr. 2005.

\bibitem{ollivier} Y. Ollivier, ``Ricci curvature of Markov chains on metric spaces,'' \textit{J. Funct. Anal.}, vol. 256, no. 3, pp. 810--864, Feb. 2009.

\bibitem{forman} R. Forman, ``Bochner's method for cell complexes and combinatorial Ricci curvature,'' \textit{Discrete Comput. Geom.}, vol. 29, pp. 323--374, 2003.

\bibitem{sreejith} R. P. Sreejith, K. Mohanraj, J. Jost, E. Saucan, and A. Samal, ``Forman curvature for complex networks,'' \textit{J. Stat. Mech.: Theory Exp.}, vol. 2016, p. 063206, Jun. 2016.

\bibitem{wee_oprc} J. Wee and K. Xia, ``Ollivier persistent Ricci curvature-based machine learning for the protein-ligand binding affinity prediction,'' \textit{J. Chem. Inf. Model.}, vol. 61, no. 4, pp. 1617--1626, Apr. 2021.

\bibitem{wee_fprc} J. Wee and K. Xia, ``Forman persistent Ricci curvature (FPRC)-based machine learning models for protein-ligand binding affinity prediction,'' \textit{Brief. Bioinform.}, vol. 22, no. 6, p. bbab136, Nov. 2021.

\bibitem{ricci_community} C.-C. Ni, Y.-Y. Lin, F. Luo, and J. Gao, ``Community detection on networks with Ricci flow,'' \textit{Sci. Rep.}, vol. 9, p. 9984, Jul. 2019.

\bibitem{sandhu} R. Sandhu, T. Georgiou, E. Reznik, L. Zhu, I. Kolesov, Y. Senbabaoglu, and A. Tannenbaum, ``Graph curvature for differentiating cancer networks,'' \textit{Sci. Rep.}, vol. 5, p. 12323, Jul. 2015.

\bibitem{cang_wei} Z. Cang and G.-W. Wei, ``TopologyNet: Topology based deep convolutional and multi-task neural networks for biomolecular property predictions,'' \textit{PLoS Comput. Biol.}, vol. 13, no. 7, p. e1005690, Jul. 2017.

\bibitem{ph_protein_wei} K. Xia and G.-W. Wei, ``Persistent homology analysis of protein structure, flexibility, and folding,'' \textit{Int. J. Numer. Methods Biomed. Eng.}, vol. 30, no. 8, pp. 814--844, Jun. 2014.

\bibitem{gudhi} The GUDHI Project, ``GUDHI User and Reference Manual,'' GUDHI Editorial Board, 2024. [Online]. Available: https://gudhi.inria.fr/doc/latest/

\bibitem{persim} H. Adams, T. Emerson, M. Kirby, R. Neville, C. Peterson, P. Shipman, S. Chepushtanova, E. Hanson, F. Motta, and L. Ziegelmeier, ``Persistence images: A stable vector representation of persistent homology,'' \textit{J. Mach. Learn. Res.}, vol. 18, no. 8, pp. 1--35, 2017.

\bibitem{pot} R. Flamary, N. Courty, A. Gramfort, M. Z. Alaya, A. Boisbunon, S. Chambon, L. Chapel, A. Corenflos, K. Fatras, N. Fournier, L. Gautheron, N. T. H. Gayraud, H. Janati, A. Rakotomamonjy, I. Redko, A. Rolet, A. Schutz, V. Seguy, D. J. Sutherland, R. Tavenard, A. Tong, and T. Vayer, ``POT: Python Optimal Transport,'' \textit{J. Mach. Learn. Res.}, vol. 22, no. 78, pp. 1--8, 2021.

\bibitem{xgboost} T. Chen and C. Guestrin, ``XGBoost: A scalable tree boosting system,'' in \textit{Proc. 22nd ACM SIGKDD Int. Conf. Knowl. Discov. Data Min. (KDD)}, San Francisco, CA, USA, Aug. 2016, pp. 785--794.
\end{thebibliography}
\end{document}